\documentclass{article}
\usepackage{ijcai23}
\usepackage{comment}
\usepackage{tabularx}
\usepackage{times}
\usepackage{soul}
\usepackage{url}
\usepackage[hidelinks]{hyperref}
\usepackage[utf8]{inputenc}
\usepackage[small]{caption}
\usepackage{graphicx}
\usepackage{amsmath}
\usepackage{amsthm}
\usepackage{booktabs}
\usepackage{algorithm}
\usepackage{algorithmic}
\usepackage[switch]{lineno}
\usepackage{amsthm,amsmath,amssymb}
\usepackage{multirow}
\DeclareUnicodeCharacter{202F}{,}

\title{Reconstruction-Aware Prior Distillation for Semi-supervised Point Cloud Completion}

\author{
Zhaoxin Fan$^{1}$ \quad \and Yulin He$^{1}$ \quad \and Zhicheng Wang$^3$ \quad \and Kejian Wu$^3$ \quad \and Hongyan Liu$^2$ \quad \and Jun He$^{1}$
\affiliations
$^1$Renmin University of China \\
$^2$Tsinghua University \\
$^3$Nreal \\
\emails
{fanzhaoxin,yolanehe,hejun}@ruc.edu.cn, hyliu@tsinghua.edu.cn, {zcwang,kejian}@nreal.ai
}

\begin{document}
\maketitle
\begin{abstract}
  Real-world sensors often produce incomplete, irregular, and noisy point clouds, making point cloud completion increasingly important. However, most existing completion methods rely on large paired datasets for training, which is labor-intensive. This paper proposes RaPD, a novel semi-supervised point cloud completion method that reduces the need for paired datasets. RaPD utilizes a two-stage training scheme, where a deep semantic prior is learned in stage 1 from unpaired complete and incomplete point clouds, and a semi-supervised prior distillation process is introduced in stage 2 to train a completion network using only a small number of paired samples. Additionally, a self-supervised completion module is introduced to improve performance using unpaired incomplete point clouds. Experiments on multiple datasets show that RaPD outperforms previous methods in both homologous and heterologous scenarios.

\end{abstract}

\section{Introduction}

Point cloud can represent the 3D geometric details of a scene or an object well with high resolution, hence attracting extensive research interests and becoming increasingly important in many applications, such as autonomous driving, robotics, and augmented reality. However, point clouds scanned from real-world sensors are always incomplete, irregular, and noisy, making many algorithms designed for point loud-based applications, such as 3D object detection \cite{zhu2020ssn}, unreliable. Therefore, point cloud completion, aiming at transforming a partially observed incomplete point cloud into its complete counterpart, becomes more and more popular in recent years.
\begin{figure}[h]
  \centering
  \includegraphics[width=0.95\linewidth]{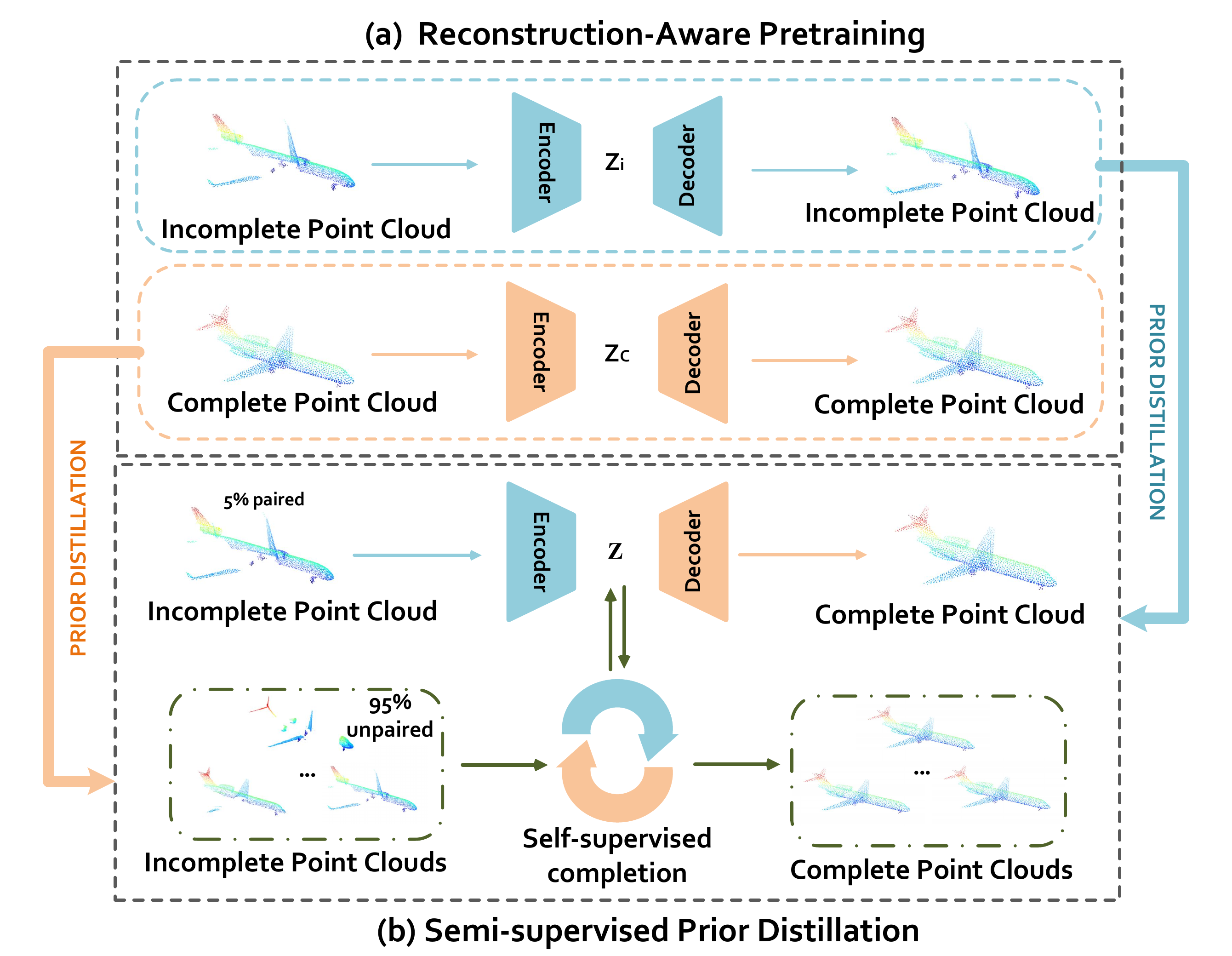}

  \caption{An illustration of our main idea. We train our point cloud completion model in two stages. In training stage 1, reconstruction-aware pretraining is introduced to learn the semantic prior. In training stage 2, the above prior is distilled into the completion network, and a self-supervised completion module is proposed to further improve the network's power. In our work, only a small number of paired training samples are required.}
  \label{fig:task}
 
\end{figure}
PCN \cite{yuan2018pcn} is a pioneering work for deep learning-based point cloud completion, which utilizes PointNet \cite{qi2017pointnet} to learn global feature of the incomplete point cloud. Then, point-wise features of the complete point cloud are folded \cite{yang2018foldingnet} from the global feature to generate the final complete point cloud.  Due to its disability in extracting local feature, PCN fails to recover the geometric details of the object. Therefore, following works \cite{liu2020morphing,pan2020ecg,wang2020cascaded,xie2020grnet,yan2022fbnet,yu2023adapointr,cai2022learning,wang2022softpool++} provide better completion results by preserving observed geometric details from the incomplete point shape using local features \cite{pan2021variational}. Though existing methods have achieved very promising performance, most of them require taking a large number of well-aligned complete-incomplete point cloud pairs to train their models, which is labor exhausted. In some wild scenarios, annotating the data is nearly impossible. Therefore, a question arises: \emph{can we train a  point cloud completion model using less paired data?}

The answer is absolutely yes. In this paper, we propose a novel deep learning model named  RaPD for semi-supervised point cloud completion. RaPD achieves the point cloud completion goal by utilizing a small number of paired point cloud samples compared to previous methods. Hence, it is much more annotation friendly and gains a significant reduction in annotation cost. The intuition behind our method is that though paired samples are hard to collect, unpaired point clouds are easy to access. For example, manually designed clean complete point clouds from shape analysis datasets like ShapeNet \cite{chang2015shapenet}, and noisy incomplete point clouds  from autonomous driving datasets like KITTI  \cite{geiger2013vision}.  The semantic prior of a category of objects consisting in these unpaired point clouds is very available and we should take full advantage of it. To this end, we propose a two-stage training scheme for RaPD to achieve the goal. Fig. \ref{fig:task}  illustrates the main idea of our method.

In training stage 1, a reconstruction-aware pretraining process is introduced. The insight behind this design is that we believe that completion-dependent features and object-shape-dependent features can be decoupled and obtained by reconstructing a point cloud using different deep learning-based models. To this end, two auto-encoders are trained for complete and incomplete point clouds separately.  The one trained using unpaired complete point clouds would characterize the completion-dependent features into its decoder and the latent global code $Z_c$, while the other one trained using unpaired incomplete point clouds would characterize the object-shape-dependent features into its encoder and  latent global code $Z_i$. The above-mentioned encoder, decoder, and latent code $Z_c$ are regarded as the semantic prior to the next training stage.

Then, in training stage 2, we distill the learned semantic prior into a point cloud completion network to enable semi-supervised training. Specifically, weights of the above-learned encoder and decoder are adopted as a good initiation. Then, object-shape-dependent features and completion-dependent features are utilized as a balanced signal to guide the network training process. Thereby, taking advantages of this kind of prior, the network can learn the ability of competing a point cloud very well utilizing only a small number of paired training samples (10\% or below than full-supervised methods in this paper). What's more, to make usage of a large amount of unpaired incomplete point clouds, we further propose a self-supervised learning module. This module would benefit from a novel category-level latent prior and an adversarial training strategy, which further increases the performance of RaPD.

To the best of our knowledge, we are the first work to design semi-supervised models for the point cloud completion task. to verify the effectiveness of our method, extensive experiments are conducted on several widely used datasets.  Experimental results show that our method can achieve excellent performance in both homologous and heterologous scenarios. When only a small number of paired training samples are available, RaPD achieves state-of-the-art performance.  And our method even achieves comparable performance to fully-supervised methods. Our contribution can be summarized as: 1) We propose the first semi-supervised point cloud completion method RaPD, which achieves state-of-the-art performance in the data shortage setting. 2) We propose a two-stage training scheme, which consists of a reconstruction-aware pretraining process and a semi-supervised prior distillation  process, enabling effective semi-supervised training. 3) We conduct extensive experiments to verify the effectiveness of our method in both homologous and heterologous scenarios. Contributions of each key design are also well-studied.

\section{Related Work}

\subsection{Point Cloud Completion}
Point cloud completion has obtained significant development in recent years. It plays an important role in autonomous driving, robotic grasping, and computer graphics. PCN \cite{yuan2018pcn} pioneers the work of deep learning-based point cloud completion. It proposes a simple encoder-decoder network for dense and complete point set generation. Since only global features are utilized, PCN is not good at  recovering the geometric details. Therefore, methods such as  \cite{sarmad2019rl,wang2020cascaded,tchapmi2019topnet,huang2020pf} are proposed to  learn local features better. For example,  Wang et al. \cite{wang2020cascaded} propose a coarse-to-fine strategy to synthesize the detailed object shapes by considering the local details of partial input with the global shape information together.  There are also some methods \cite{wang2020softpoolnet,zhang2020detail,xie2020grnet}  propose to learn better semantics for generating more complicated and cleaner complete point clouds. For example, SoftPoolNet \cite{wang2020softpoolnet} proposes a soft pooling operation to adaptively extract better semantics features for point cloud generation. To learn finer details, VRCNet \cite{pan2021variational} proposes a variational framework that consists of a probabilistic modeling module and a relational enhancement module to improve performance. Though promising, these methods are limited in that they all require a large number of paired complete-incomplete point cloud pairs for training, which is cost unfriendly. To tackle the issue, Zhang et al. \cite{zhang2021unsupervised} propose to use GAN-Inversion \cite{xia2021gan} for unsupervised point cloud completion. Though unpaired samples are no longer required, time-consuming inference time optimization is needed during testing in \cite{zhang2021unsupervised}, making it in-efficient. Besides,  in \cite{zhang2021unsupervised}, one model only works for a particular category, further limiting its broader application.  Most similar to our work is \cite{chen2019unpaired}, which also use an encoder-decoder structure for point cloud completion.  Howecer, firstly, our approach is semi-supervised, whereas  \cite{chen2019unpaired} is not. Secondly, we use a discriminator on incomplete point clouds with the help of a degradation module, while
the  \cite{chen2019unpaired} uses a GAN for mapping two spaces
on the completion point cloud. Thirdly, we propose a semisupervised prior distillation technique that distills different features as prior, which is not discussed in t \cite{chen2019unpaired}. Finally, we address the case when pair-data is limited, which is not discussed in  \cite{chen2019unpaired}.

In contrast to all previous works, we propose a novel reconstruction-aware prior distillation method named RaPD for semi-supervised point cloud completion. To tackle the data shortage problem, RaPD takes full advantage of the semantic prior hidden in unpaired point clouds and distills the learned prior into the final completion network. Our method achieves superior performances than both previous supervised and unsupervised methods only taking a small number of paired point clouds for training.
\begin{figure*}[h]
  \centering
  \includegraphics[width=0.95\linewidth]{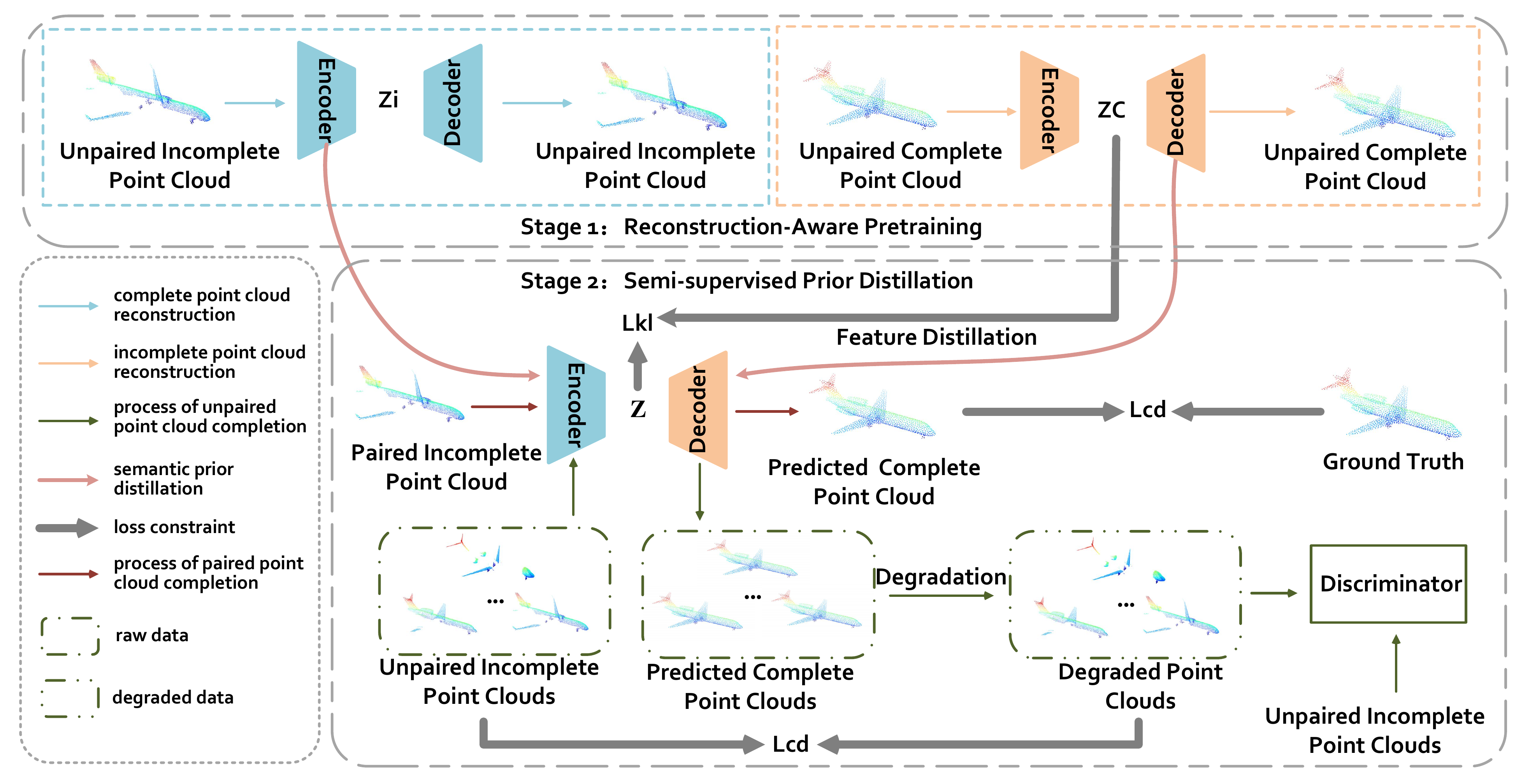}
 
  \caption{Framework of RaPD. Our framework adopts a two stage training scheme: the reconstruction-aware pretraining stage and the semi-supervised prior distillation stage. The former learns high-quality semantic prior, while the later learns the final completion network by distilling the semantic prior.}
  \label{fig:framework}

\end{figure*}
\subsection{Semi-supervised Learning}
Semi-supervised learning is a mixed research task between supervised and unsupervised tasks \cite{chapelle2009semi,zhu2005semi}. It targets predicting more accurately with the aid of unlabeled data than supervised learning that uses only labeled data \cite{yang2021survey}. Semi-supervised methods can be divided into generative methods \cite{denton2016semi,dai2017good}, consistency regularization methods \cite{rasmus2015semi,sajjadi2016regularization}, graph-based methods \cite{iscen2019label,chen2020data}, pseudo-labelling methods \cite{grandvalet2004semi,lee2013pseudo}, and hybrid methods \cite{berthelot2019mixmatch,berthelot2019remixmatch}. With the development of deep learning, semi-supervised learning has been deployed into a wide range of downstream tasks. For example, Tang et al. \cite{tang2016large} propose a semi-supervised object detection method using visual and semantic knowledge transfer. Zhou et al. \cite{zhou2021r} introduce a recurrent multi-scale feature modulation approach for monocular depth estimation. Quali et al. \cite{ouali2020semi} present a method that utilizes cross-consistency training for semi-supervised semantic segmentation. In the field of point cloud processing, Huang et al. \cite{huang2020feature} propose a fast and robust semi-supervised approach for point cloud registration. Zhao et at. \cite{zhao2020sess} propose a self-ensembling technique in their work for semi-supervised 3D object detection.  Cheng et al. \cite{cheng2021sspc} propose to divide a point cloud into superpoints and build superpoint graphs for semi-supervised point cloud segmentation. Though semi-supervised learning has attracted massive attention, to our knowledge, there is no existing method tailored for point cloud completion. 

In this paper, we propose RaPD, which is tailored for the point cloud completion task. In our work, discriminative semantic prior hidden in unpaired point clouds are distilled and utilized to achieve our semi-supervised training goal.

\section{Method}

\subsection{Overview}

\noindent \textbf{Problem Statement:} We aim at training a deep learning model that takes a incomplete point cloud $P_i$ as input, it can output  $P_i$'s complete counterpart  $P_c$. In this paper, in order to reduce the dependence on paired complete-incomplete samples,  we study the semi-supervised point cloud completion task. In our setting, there exists two training datasets $\mathcal{D}_{paired}=\{(P_{i}^1,P_{c}^1), (P_{i}^2, P_{c}^2), \cdots ,(P_{i}^K,P_{c}^K)\}$ and  $\mathcal{D}_{unpaired} =\{P_{i}^1, P_{i}^2,\cdots, P_{i}^N, P_{c}^1, P_{c}^2, \cdots, P_{c}^M\}$, where $K$, $N$ and  $M$ are number of point clouds. Point clouds in $\mathcal{D}_{paired}$ are well-paired that can be utilized for supervised training, while point clouds in $\mathcal{D}_{unpaired}$  are unpaired samples with great diversity. A constrain between the two datasets is that $K \textless\textless N(M) $, Which means $\mathcal{D}_{paired}$ is small scale. The deep learning model should be trained  utilizing $\mathcal{D}_{paired}$  and $\mathcal{D}_{unpaired}$. The expectation is that the final trained model can do point cloud completion excellently even though only a small number of training samples are labeled (paired).
 
\noindent \textbf{Framework of the Approach:} To solve the semi-supervised point cloud completion problem, we propose a novel method named RaPD. Fig. \ref{fig:framework} illustrates the framework. RaPD adopts a two-stage training scheme. In the first stage, we conduct reconstruction-aware pretraining to learn the semantic prior from $\mathcal{D}_{unpaired}$. Specifically, two auto-encoders are trained using the complete point clouds and incomplete point clouds, respectively. After trained, network weights and embedded  latent codes of point clouds are regarded as the semantic prior, which would be exploited in the next stage. Then, in the second stage, we conduct semi-supervised prior distillation. This stage consists of a prior distillation module and a self-supervised completion module, both of which would take  $\mathcal{D}_{unpaired}$ and $\mathcal{D}_{paired}$ as training data and jointly learn an excellent model for the final point cloud completion task. 

\subsection{Stage 1: Reconstruction-Aware Pretraining}
Since it is very hard to annotate complete-incomplete point cloud pairs in real-world scenarios, only a small number of paired point clouds ($K$ in $\mathcal{D}_{paired}$) can be leveraged to train a model in the semi-supervised training setting. However, we argue that though paired data is rare, there are abundant unpaired point clouds that can be leveraged. For complete point clouds, there exist human-designed object datasets such as ShapeNet \cite{chang2015shapenet}. For incomplete point clouds, we can extract samples from the wild datasets such as KITTI \cite{geiger2013vision}. They consist of  $\mathcal{D}_{unpaired}$. We note here that the semantic characteristics in these unpaired point clouds are very valuable and should be exploited to benefit the semi-supervised point cloud completion task. Therefore, we propose to learn the semantic prior from them. In fact, there are two kinds of characteristics that would benefit our task: the completeness information of an object and the shape information of an object. We term the two characteristics as completion-dependent features and object-shape-dependent
features respectively. And we find that reconstruction-aware pretraining can learn them effectively.

\noindent \textbf{Learning completion-dependent features}: To learn the completion-dependent features, we train an auto-encoder using the $M$ complete point clouds in  $\mathcal{D}_{unpaired}$. As shown in Fig. \ref{fig:framework}, the auto-encoder consists of an encoder and a decoder. Taking a complete point cloud $P_c$ as input, the encoder would embed it into a latent code $Z_c$. Then, the decoder would take $Z_c$ as input and recover the shape of $P_c$, termed $\hat{P_{c}}$. After training, the latent code $Z_c$ would contain the completeness information of  $P_c$ since  $\hat{P_{c}}$ is recovered only utilizing $Z_c$, the predicting of which is learned from massive complete point clouds without disturbed by any incomplete noise.  Besides, since the decoder $D_c$ works for recovering a completion point cloud, the network weights of the decoder would also contain rich implicit completeness-related information. Therefore, the decoder and  $Z_c$ together  consist of the completion-dependent features. And  we regard the latent code $Z_c$ and the decoder  as part of the expected semantic prior.

\noindent \textbf{Learning Object-Shape-Dependent Features}: The object-shape-dependent features are learned in the similar way. The difference is that the auto-encoder is trained by using the $N$ incomplete point clouds in $\mathcal{D}_{unpaired}$ . Similarly, after training, the encoder $E_i$ and latent code $Z_i$ of the input point cloud would be very expressive in reflecting the shape of the object in an implicit way, i.e, they are actually the object-shape-dependent features that we expect to learn. It is also very straightforward to regard the two as part of the semantic prior. However, since  $Z_i$ is generated by $E_i$, they are essentially redundant. Therefore, we only take the encoder $E_i$ as the semantic prior.  In another word, The encoder $E_i$ is used to encode the input point cloud into a compact and low-dimensional feature representation, which contains the object-shape-dependent features. These features capture the overall shape information of the object, which we believe is crucial for accurate 3D point cloud completion.

\subsection{Stage 2: Semi-supervised Prior Distillation}
In this stage, we focus on training an encoder-decoder network to achieve the final point cloud completion goal taking $\mathcal{D}_{paired}$, $\mathcal{D}_{unpaired}$ and the semantic prior as training data. To do this, we design two modules. The first module is the distillation module, which mainly utilizes  $\mathcal{D}_{paired}$ as input and focuses on distilling information from the semantic prior to better train the model. The second module is the self-supervised completion module, which mainly makes usage of the incomplete point clouds in $\mathcal{D}_{unpaired}$ to further benefit the network.

\noindent \textbf{Prior Distillation:} The distillation module quantifies the semantic prior into two categories. The first category is named weights distillation. In particular, we build the encoder-decoder network architecture to be the same as the auto-encoders in training stage 1. And we use the weights of the trained  $E_i$ and $D_c$ as the initialization of the encoder $E$ and decoder $D$. Since weights of $E_i$ and $D_c$ are equipped with the ability of representing the shape-related information and completeness-related information, such a kind of distillation would flow very valuable prior information from training stage 1 to training stage 2. Besides, since $E_i$ and $D_c$ are already well capable of point cloud reconstruction, initiating the network using them would help the model coverage faster, further increasing the robustness of the model and avoiding the network from falling into local-minimal.

Besides distilling model weights, we also propose a  features distillation step.  Specifically, one observation is that since we initialize the decoder using $D_c$, if the latent code $Z$ was the same as or at least be very similar to $Z_c$, the decoder can nearly perfectly output a complete point cloud. The reason is natural that $D_c$ is already trained for recovering a complete point cloud in stage 1 and can perform well if the latent code is sampled from the feature space of $Z_c$. Therefore, to promote the model to learn such a kind of prior, we encourage the distribution of $Z$ to be close to that of $Z_c$ during training. To achieve so, we adopt the Kullback-Leibler divergence as a loss function:

\begin{equation}
L_{z,paired}=D_{kl}(Z_c||Z)
\end{equation}

Then, for point cloud in $\mathcal{D}_{paired}$, we use the chamfer distance loss as the point cloud similarity loss.

\begin{align}
L_{cd,paired} &= \frac{\gamma}{rN} \sum_{x \in P_i} \min_{y \in P_c}||x-y||_2^2 \notag \\
&\phantom{=} + \frac{1}{rN} \sum_{y \in P_c} \min_{x \in P_i}||y-x||_2^2
\end{align}

where $\gamma$ is a balance term.
	\begin{table*}
\begin{tabular}{c|cc|cc|cc|cc}
\hline
           & \multicolumn{ 2}{|c}{1\%} & \multicolumn{ 2}{|c}{2\%} & \multicolumn{ 2}{|c}{5\%} & \multicolumn{ 2}{|c}{10\%} \\
\cline{2-9}
            &         CD &   F1-score &         CD &   F1-score &         CD &   F1-score &         CD &   F1-score \\
\hline
    TopNet \cite{tchapmi2019topnet}&      20.73 &      0.186 &      16.49 &       0.220 &      14.37 &      0.236 &      13.62 &      0.251 \\

       PCN \cite{yuan2018pcn} &       20.80 &      0.199 &      16.16 &      0.229 &      12.52 &      0.272 &      10.99 &      0.299 \\

    VRCNet \cite{pan2021variational} &      30.65 &      0.214 &      20.36 &      0.252 &      13.36 &      0.319 &       \textbf{9.38} &      \textbf{0.415} \\

    RaPD*  &      15.19 &      0.263 &       12.80 &      0.287 &       10.90 &      0.315 &      10.63 &      0.325 \\

   RaPD &      \textbf{12.12} &      \textbf{0.313} &      \textbf{11.07} &      \textbf{0.319} &      \textbf{10.07} &      \textbf{0.328} &       9.64 &      0.335 \\
\hline
\end{tabular} 

\caption{Comparison on the MVP dataset under the semi-supervised training setting. CD is loss is multiplied by $10^4$. 1\% to 10\% are the percentages of paired point clouds in the training set that are used for training.} 
\label{tab:mvp_semi}
\end{table*}

\begin{table*}
\centering
\begin{tabular}{c|c|cc}
\hline
           &    Methods        &      CD &   F1-score \\
\hline
\multicolumn{ 1}{c|}{fully supervised} &        PCN \cite{yuan2018pcn} &       9.77 &       0.320 \\

\multicolumn{ 1}{c|}{} &     TopNet \cite{tchapmi2019topnet} &      10.11 &      0.308 \\

\multicolumn{ 1}{c|}{} &        MSN \cite{liu2020morphing} &        7.90 &      0.432 \\

\multicolumn{ 1}{c|}{} &        ECG \cite{pan2020ecg} &       6.64 &      0.476 \\

\multicolumn{ 1}{c|}{} & Wang et al. \cite{wang2020cascaded} &       7.25 &      0.434 \\

\multicolumn{ 1}{c|}{} &     VRCNet \cite{pan2021variational} &       \textbf{5.96} &      \textbf{0.499} \\
\hline
\multicolumn{ 1}{c|}{semi-supervised} &      RaPD* &      10.63 &      0.325 \\

\multicolumn{ 1}{c|}{} &       RaPD &       \textbf{9.64} &      \textbf{0.335} \\
\hline
\end{tabular}  

\caption{Comparison with fully-supervised methods on the MVP dataset. CD loss is multiplied by $10^4$. RaPD and PaPD* are trained using on 10\% of paired training data, while all other methods are training using all paired training data.} 
\label{tab:mvp_fully}

\end{table*}
\noindent \textbf{Self-supervised completion:} During training, beyond using $\mathcal{D}_{paired}$,  we also take the incomplete point clouds in $\mathcal{D}_{unpaired}$ as training data. To take advantage of them, we propose a self-training strategy. Specifically, for point cloud $P_i$ in  $\mathcal{D}_{unpaired}$, it is first fed into the network to predict a point cloud $P_n$. Then, since there is no ground truth for $P_n$, we can only make use of the input $P_i$ as a training signal. To do so, we should first degrade $P_n$ to  $P_i$, termed $\hat{P_i}$. For the degradation, we adopt the K-Mask Degradation Module in \cite{zhang2021unsupervised}. Then, the loss function for this signal is set as the chamfer distance loss too: 

\begin{align}
L_{cd,unpaired} &= \frac{\gamma}{rN} \sum_{x \in P_i} \min_{y \in \hat{P_i}}||x-y||_2^2 \notag \\
&\phantom{=} + \frac{1}{rN} \sum_{y \in \hat{P_i}} \min_{x \in P_i}||y-x||_2^2
\end{align}

$L_{cd,unpaired}$ would encourage $P_n$ to be as similar as its incomplete counterpart $P_i$  as possible during training.
However, $L_{cd,unpaired}$ only supervises the similarity in a low-level manner, which is not powerful enough for the completion task. To further benefit the encoder-decoder network, we add a discriminator $\mathbb{D}$  to guide the training process. The discriminator plays the role of encouraging $\hat{P_i}$  to be as similar  $P_i$ as possible. It would predict a probability for each of its input, where close to 0 means it regards the point cloud as a fake incomplete one, while close to 1 means it regards the point cloud as a real incomplete one. In this way, the network would be forced to generate a more realistic $\hat{P_i}$ . Hence, the high-level shape semantics between $P_n$ and $P_i$ would also be aligned. The loss function for the discriminator is:

\begin{equation}
L_d=(\mathbb{D}(P_i)-1)^2+(\mathbb{D}({\hat{P_i}}))^2
\end{equation}

\begin{equation}
L_g=(\mathbb{D}({\hat{P_i}})-1)^2
\end{equation}

 Though straightforward,  only using  $L_{cd,unpaired}$  would cause over-fitting since there is no signal account for the completeness of $P_n$. To encourage $P_n$ to also be as complete as possible, we  use the following loss to make the distribution of the latent code $Z$ fall into the completeness-related space:
 
\begin{equation}
L_{z,unpaired}=D_{kl}(\hat{Z_c}||Z)
\end{equation}

where $\hat{Z_c}$ is the category-level mean latent code of all complete training point clouds in training stage 1. This loss can also be regarded as distilling a part of semantic prior from stage 1, further taking advantage of $\mathcal{D}_{unpaired}$.

For training stage 1. We train two auto-encoders to formulate the reconstruction-aware pretraining. The loss functions of training both networks are chamfer distance loss. For training stage 2, the loss function for the encoder-decoder network is the weighted sum of each component of the training process:

\begin{equation}
\begin{split}
L=\lambda_1 L_{z,paired}+ \lambda_2 L_{z,unpaired}+ \lambda_3 \\ L_{cd,paired}+ \lambda_4 L_{cd,unpaired} +\lambda_5 L_{g}
\end{split}
\end{equation}

where $\lambda_1$ to $\lambda_5$ are balance terms. For the discriminator, it is optimized using $L_d$ independently.

\section{Experiment}
\subsection{Implementation Details}
To verify the superiority of RaPD, we conduct experiments on three widely used public datasets: MVP \cite{pan2021variational}, CRN \cite{wang2020cascaded} and KITTI \cite{geiger2013vision}.  We set our encoder-decoder network architecture to be the same as PCN \cite{yuan2018pcn} for a fair comparison. The architecture of the discriminator follows the basic architecture of \cite{li2019pu}. The evaluation metrics are chamfer distance (CD) and the F1 score,  following \cite{pan2021variational}. More details please refer to the SuppMat.

\subsection{Main Results}
\begin{figure}[h]
  \centering
  \includegraphics[width=0.9\linewidth]{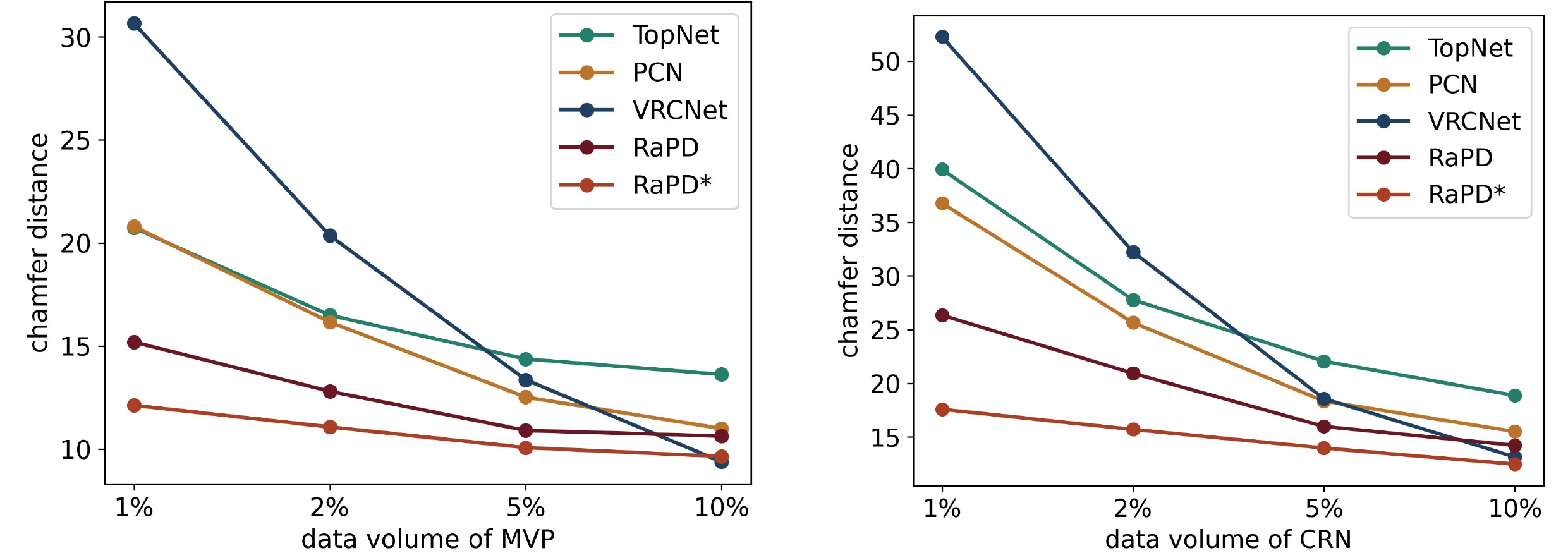}
 
  \caption{Comparison on MVP and CRN. }
  \label{fig:chart}
 
\end{figure}

\begin{figure}[h]
  \centering
  \includegraphics[width=0.8\linewidth]{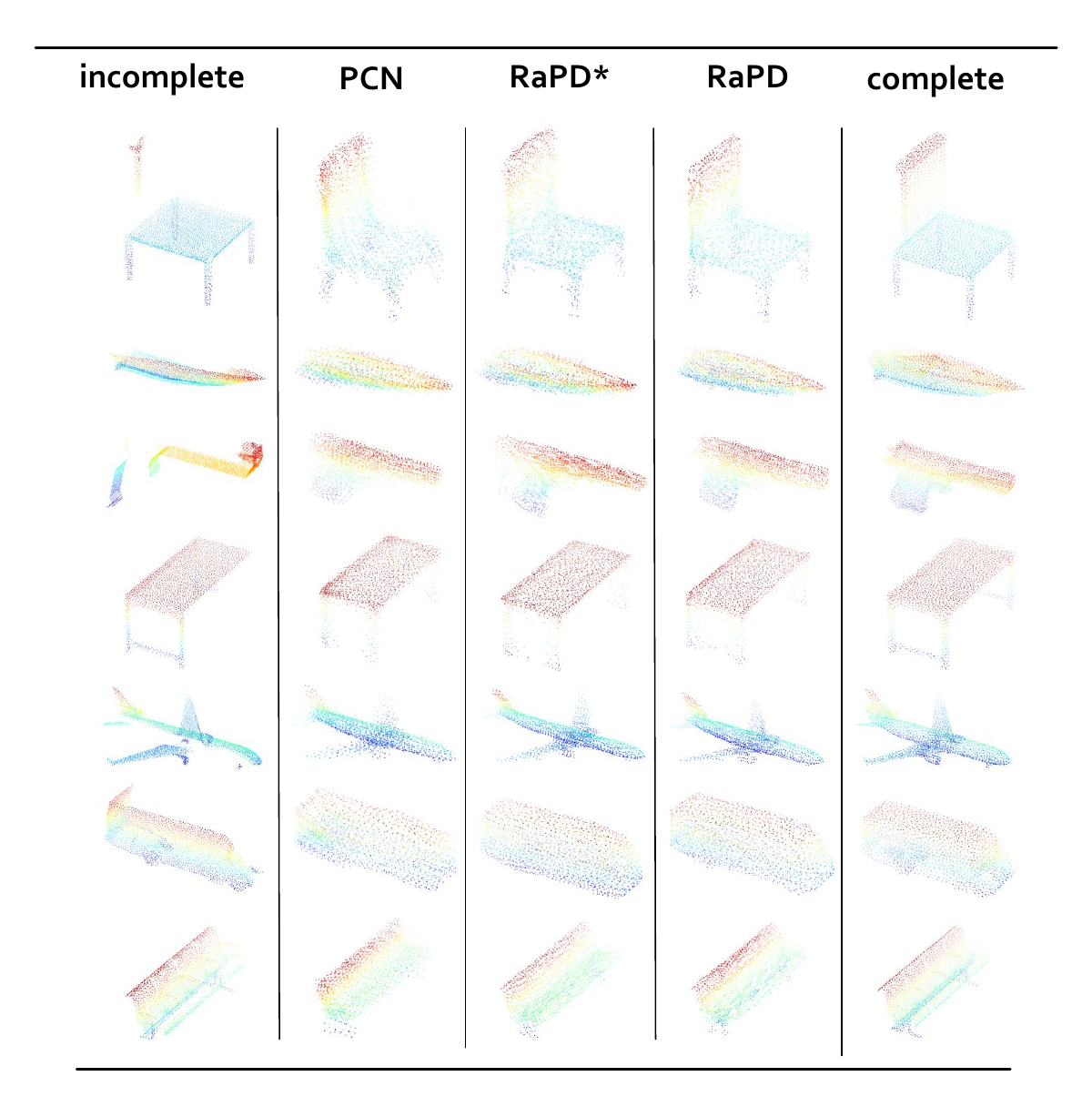}

  \caption{Quantitative comparison on MVP. The models are trained using 5\% of paired point clouds form the training set.}
  \label{fig:mvp}

\end{figure}

\noindent \textbf{Results on MVP dataset:} We first compare our method with PCN \cite{yuan2018pcn}, TopNet \cite{tchapmi2019topnet} and VRCNet \cite{pan2021variational} on the MVP dataset. Since there is no existing semi-supervised point cloud completion method, for a fair comparison, we re-train their model using the same number of paired point clouds to report their performance under our semi-supervised training setting. We train all models on the MVP training set and test them on the MVP testing set. Therefore, the training and testing datasets are homologous. Table \ref{tab:mvp_semi} and Fig. \ref{fig:chart}  show the results.  RaPD* is a version of RaPD that discard the self-supervised completion module during training. It can be seen that with the increase of paired training data, performances of all models are promoted. However, no matter whether we adopt the self-supervised completion module, our method outperforms all existing methods by a large margin. These results greatly demonstrate the advantage of our idea of distilling semantic prior from large-scale unpaired data over existing supervised methods. The performance gain of RaPD over RaPD* also demonstrates the superiority of the  self-supervised completion module in improving the performance. 
We find when using 10\% data, RaPD is outperformed by VRCNet, that is because RaPD is designed for small number of labeled data. Here 10\% data reaches around 6000 paired point clouds, which breaks the small data rule in practice, so it would benefit VRCNet more.  But note, in real world scenarios, it is hard to collect so many  paired data. Some visualization results are shown in Fig. \ref{fig:mvp}. From the visualization, it can be seen that RaPD can obviously recover better geometric details than PCN and RaPD*. For example, the leg of the complete chair predicted by RaPD is obviously more clear and more slender than the one predicted by PCN. 

We also compare RaPD with existing state-of-the-art supervised methods. All compared methods are training using all the paired data in the training set while RaPD and RaPD* are only trained using 10\% of paired data. The result is shown in Table \ref{tab:mvp_fully}. It can be seen that though we only use 10\% of paired training data, our method can achieve comparable performance with fully-supervised methods. Note that RaPD even outperforms PCN and TopNet. As mentioned before, RaPD  shares the same network architecture with PCN. However, it outperforms PCN using only a small number of data, which further demonstrates the superiority of our semi-supervised training scheme. 

\begin{figure}[h]
  \centering
  \includegraphics[width=0.8\linewidth]{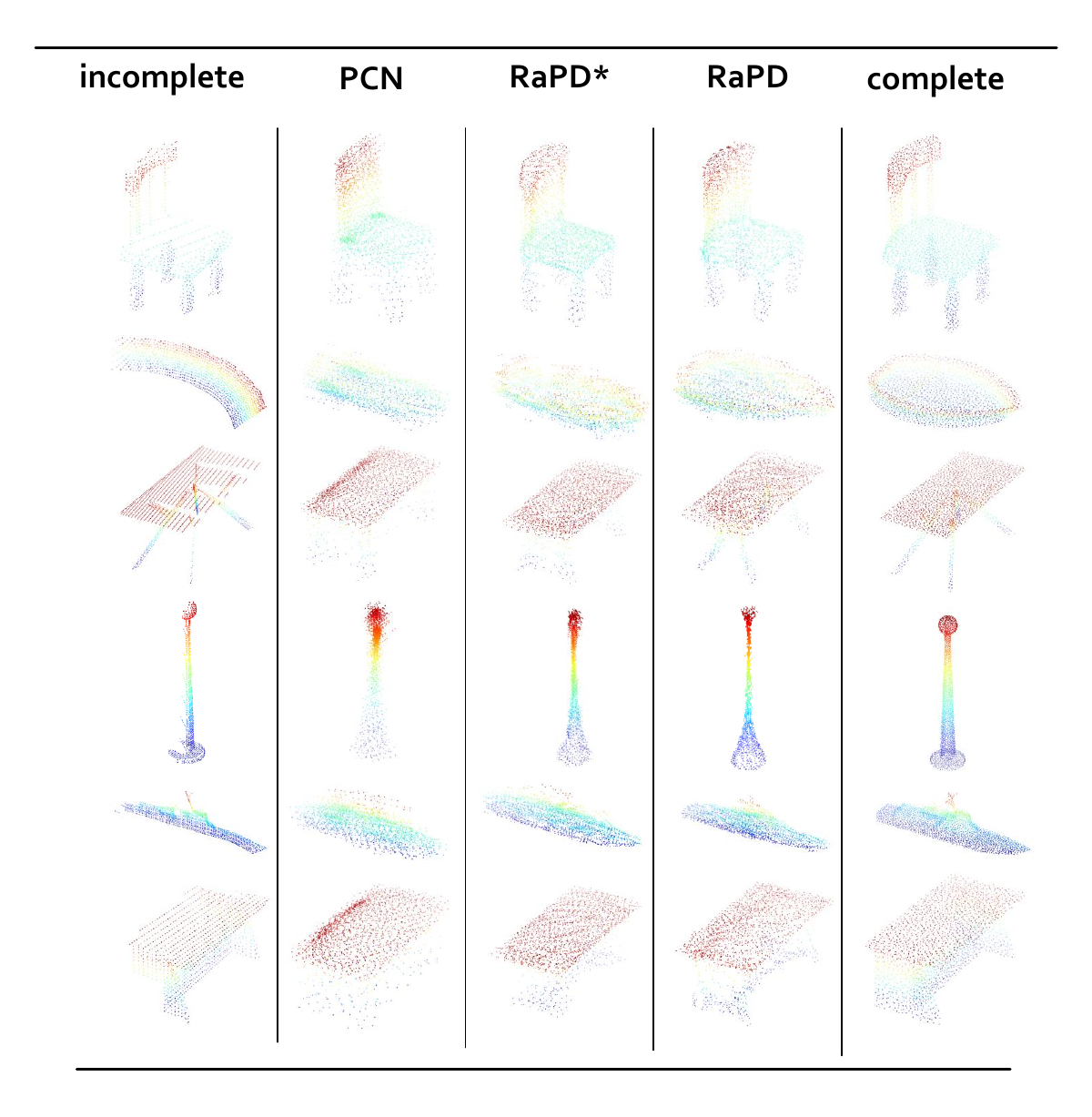}
 
  \caption{Quantitative comparison on CRN dataset.}
  \label{fig:crn}
 
\end{figure}

\begin{figure*}[h]
  \centering
  \includegraphics[width=0.9\linewidth]{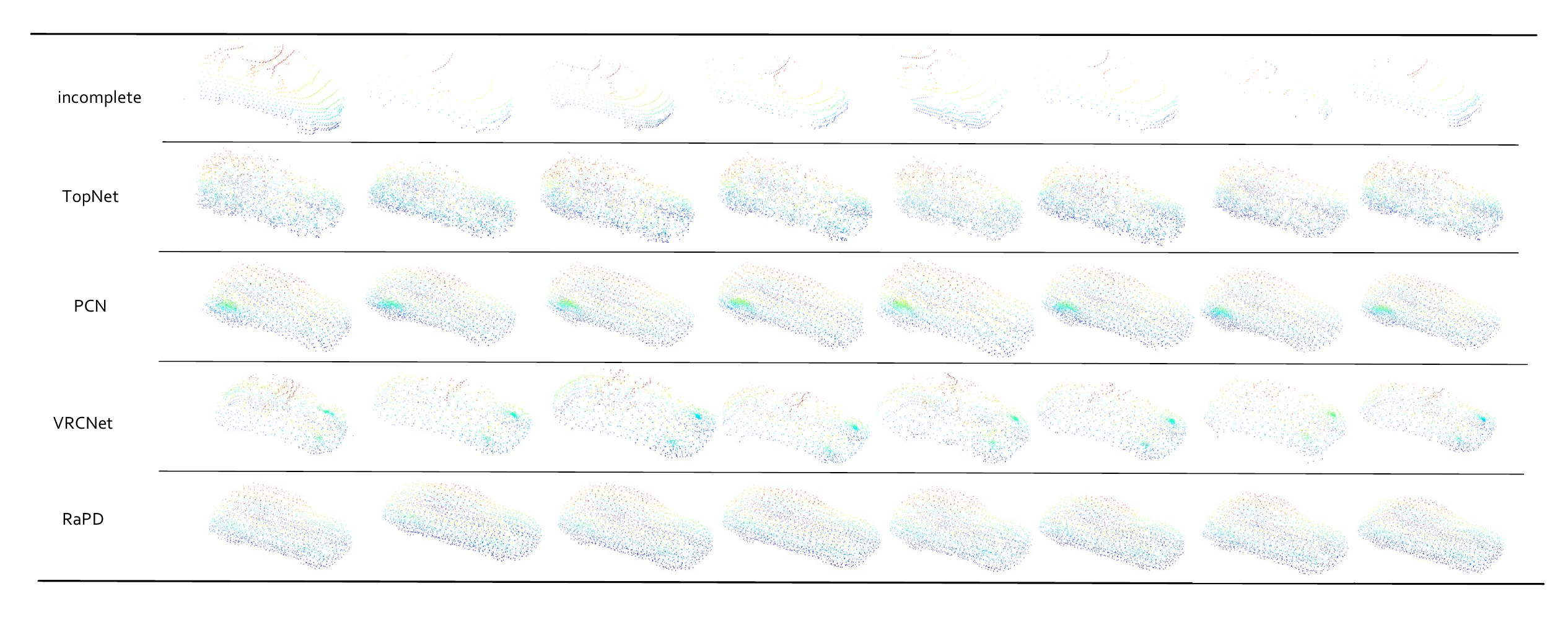}

  \caption{Quantitative comparison on KITTI dataset.}
  \label{fig:kitti}
 
\end{figure*}

 \begin{table*}
 \centering
\begin{tabular}{c|cc|cc|cc|cc}
\hline
           & \multicolumn{ 2}{|c}{1\%} & \multicolumn{ 2}{|c}{2\%} & \multicolumn{ 2}{|c}{5\%} & \multicolumn{ 2}{|c}{10\%} \\
\cline{2-9}
           &         CD &   F1-score &         CD &   F1-score &         CD &   F1-score &         CD &   F1-score \\
\hline
    TopNet \cite{tchapmi2019topnet} &      39.91 &      0.125 &      27.77 &      0.157 &      22.05 &      0.171 &      18.86 &      0.193 \\

       PCN \cite{yuan2018pcn} &      36.76 &       0.130 &      25.65 &      0.165 &      18.34 &      0.195 &       15.50 &      0.215 \\

    VRCNet \cite{pan2021variational} &      52.28 &      0.115 &      32.24 &      0.137 &      18.56 &      0.215 &      13.13 &      \textbf{0.274} \\

     RaPD* &      26.35 &      0.182 &      20.92 &      0.206 &      15.98 &      0.229 &      14.21 &      0.246 \\

      RaPD &     \textbf{17.57} &      \textbf{0.229} &      \textbf{15.70} &     \textbf{0.235} &      \textbf{13.96} &      \textbf{0.244} &      \textbf{12.47} &      0.272 \\
    \hline

\end{tabular}  

\caption{Comparison on the CRN dataset under the semi-supervised training setting. CD loss is multiplied by $10^4$. 1\% to 10\% are the percentages of paired point clouds in the training set that are used for training.} 
\label{tab:crn_semi}

\end{table*}

\begin{table}
\centering
\small
\setlength{\tabcolsep}{3.5pt}
\begin{tabular}{c|cc|cc}
\hline
           & \multicolumn{ 2}{|c|}{RaPD*} & \multicolumn{ 2}{|c}{RaPD} \\
\cline{2-5}
           &         CD &   F1-score &         CD &   F1-score \\
\hline
w/o features distillation &    11.61 &      0.299  &   43.86   &   0.109    \\

w/o decoder distillation &     11.95 &     0.283 &    10.74    &     0.321 \\

w/o encoder distillation &     11.88  &     0.292 &  10.77   &     0.315 \\

w/o weights distillation &     12.52  &     0.272 &  11.66    &     0.309 \\

full model & {\bf 10.90} & {\bf 0.315} & {\bf 10.07} & {\bf 0.328} \\
\hline
\end{tabular}

\caption{Ablation study: impact of prior distillation.} 
\label{tab:impact_distil} 

\end{table}

\noindent \textbf{Results on  CRN:} As mentioned before, the training data and testing data of training stage 1 and training stage 2 are homologous when we evaluate our method on the MVP dataset. To verify the performance of our method in the heterologous situation, we conduct experiments on CRN dataset. Specifically, we conduct the reconstruction-aware pretraining on the MVP training set and conduct the semi-supervised prior distillation on the CRN training set, and test the trained model on the CRN testing set. The results are shown in Table \ref{tab:crn_semi}, Fig. \ref{fig:chart}, and Fig. \ref{fig:crn}. It can be seen that RaPD* and RaPD significantly outperform the competitors,  benefited from our prior distillation design and the semi-supervised completion module.  This experiment demonstrates that our method not only performs well in  homologous scenario, but also generates good generalization ability in heterologous scenario. That is, when generalizing to a new domain, the model only needs fine-tune or retrain in a semi-supervised prior distillation way based on the learned semantic prior to achieve an excellent performance.

\noindent \textbf{Results on KITTI:} We also compare our method with TopNet \cite{tchapmi2019topnet}, PCN \cite{yuan2018pcn}, and VRCNet \cite{pan2021variational} in the wild KITTI dataset. Since there is no complete ground truth for objects in the KITTI dataset, we only show the qualitative comparison in Fig. \ref{fig:kitti}. It can be seen that the complete point clouds generated by our method are much better than the mentioned three methods. For example, the rear of the car is more smooth than that of PCN, TopNet, and VRCNet. The results demonstrate that RaPD can generalize well in wide scenarios. With less training overhead and better performance, RaPD has the potential of benefiting downstream applications like autonomous driving.

\begin{table}
\centering
\begin{tabular}{c|cc}
\hline
\multicolumn{ 1}{c|}{} & \multicolumn{ 2}{|c}{RaPD} \\
\cline{2-3}
\multicolumn{ 1}{c|}{} &         CD &   F1-score \\
\hline
w/o discriminator &      11.83 &      0.307 \\

full model & {\bf 10.07} & {\bf 0.328} \\
\hline
\end{tabular}  

\caption{Ablation study: impact of the discriminator.}
\label{tab:impact_discriminator}

\end{table}

\subsection{Ablation Study}
In this section, we conduct extensive ablation study to verify the effectiveness of our key designs. All experiments are conducted on the MVP dataset. 5\% paired training samples are used for training. More ablation study can be found in the SuppMat.

\noindent \textbf{Impact of Prior Distillation:} In our work, we present two kinds of prior distillation for the final completion network: the weights distillation (including encoder weights distillation and decoder weights distillation) and the features distillation. The two kinds of distillations are both one of the most critical factors for the success of our semi-supervised point cloud completion network. Table \ref{tab:impact_distil} shows the impact of them.  For features distillation, when it is discarded, the performance drops a lot, which demonstrates that it is very necessary to encourage the distribution of latent code of the input incomplete point cloud to be similar to that of the large scale unpaired complete point clouds. Then, we find that both the encoder weights distillation and decoder weights distillation are beneficial to our completion network. Since any of them is removed, the performance will significantly drop.  When both of them are removed, the performance drop is even more serious. In a world, it can be found that both kinds of distillation play a significant role in ensuring the final completion performance. The absence of any of them would lead to a sharp performance decrease.
\begin{table}
\centering
\begin{tabular}{c|cc}
\hline
           & \multicolumn{ 2}{|c}{RaPD} \\
\cline{2-3}
           &         CD &   F1-score \\
\hline
random downsample &  82.29    &   0.037         \\

voxel mask &     10.81       &     0.312       \\

  tau mask &   14.46         &   0.311         \\

    k mask & {\bf 10.07} & {\bf 0.328} \\
\hline
\end{tabular}  

\caption{Ablation study: impact of degradation methods.}
\label{tab:impact_degradation}

\end{table}
\noindent \textbf{Impact of Self-Supervised Completion:} In our work, we introduce a self-supervised completion module to make use of the abundant unpaired incomplete point clouds in training stage 2. The performance difference between the RaPD* and RaPD has demonstrated that this module is very important for performance gain since it helps the model learn more information from a large scale of incomplete unpaired point clouds in an effective way.  In this module, we use a degradation method to get a degraded point cloud for loss computation. We find that the choice of the degradation method we use is one of the key components that may affect the performance. In Table \ref{tab:impact_degradation}, we study the impact of several different degradation methods. It can be found that the K-Mask method is the best choice. The reason may be that K-Mask can better preserve the geometric details of point cloud after degradation. What's more, we use a discriminator for adversarial training to further increase the realism of the predicted point cloud. In  Table \ref{tab:impact_discriminator}, we illustrate the performance of discarding the discriminator and the corresponding adversarial loss. It can be found that without them, the performance drops a lot. Therefore, it is necessary to adopt the discriminator during training.

\section{Conclusion}
Point cloud completion is an important but challenging task for many real-world applications. To reduce the dependency on paired data when training a point cloud completion model, this paper proposes a novel semi-supervised method named RaPD. A two-stage training scheme is introduced in RaPD. The first stage learns powerful semantic prior through reconstruction-aware pretraining, while the second stage learns the final completion deep model by prior distillation and self-supervised completion learning. Experiments on several widely used datasets have demonstrated the superiority of RaPD over previous supervised methods. In the future, we will research efficient unsupervised point cloud completion to further ease the training requirement.

\section*{Acknowledgements}
This work was supported in part by the National Key Research and Development Program of China under Grant No. 2020YFB2104101 and the National Natural Science Foundation of China (NSFC) under Grant Nos. 62172421, and 62072459.

\section*{Contribution Statement}
Zhaoxin Fan and Yulin He contributed equally to this work. Jun He is the corresponding author.

\section*{Appendix}
\subsection{Dataset details}
To verify the superiority of RaPD, we conduct experiments on three widely used public datasets: MVP \cite{pan2021variational}, CRN \cite{wang2020cascaded} and KITTI \cite{geiger2013vision}.

\noindent \textbf{MVP} is a high-quality multi-view partial point cloud dataset. The point clouds are sampled from the well-known ShapeNet \cite{chang2015shapenet} benchmark. In MVP, there are a total of 16 categories. Its official splitting contains 62400 point cloud pairs for training and 41600  pairs for testing.

\noindent \textbf{CRN} is also a popular dataset constructed by sampling point cloud from ShapeNet benchmark. It consists of 8 categories. Incomplete counterparts of complete point clouds are generated by first rendering the complete point clouds into 2.5D depth maps and then back-projecting them into point clouds. CRN contains 28974 point cloud pairs for training and  800 pairs for testing.

\noindent \textbf{KITTI} is a large-scale real-world dataset originally constructed for autonomous driving. We follow \cite{yuan2018pcn} to use the ground-truth 3D bounding box annotation to extract the partially observed objects from each scan. The extracted point clouds are used to test the generalization ability of the model.

\subsection{Implementation Details}

We implement our method using the PyTorch \cite{paszke2017automatic} platform. We set our encoder-decoder network architecture to be the same as PCN \cite{yuan2018pcn} for a fair comparison. The architecture of the discriminator follows the basic architecture of \cite{li2019pu}, which consists of a global feature extractor, attention units, and a set of MLPs for final confidence regression. Both the number of points of the input and output point clouds are set to 2048.  The dimension of the latent code is 1024. $\lambda_3$ to $\lambda_5$  are set to 1, 0.5 and 0.1 respectively. $\lambda_1$ and $\lambda_2$ are both set as 5, 2, 1 at the 1st, 20th and 100th epoch respectively. We optimize our network work using the Adam optimizer \cite{kingma2014adam} with a batch size of 32.  The initial learning rate is 0.0001. In the Reconstruction-Aware Pretraining stage, the two auto-encoders are trained for 300 epochs. In the Semi-supervised Prior Distillation stage, the network is trained for 150 epochs.  The evaluation metrics are chamfer distance (CD) and the F1 score, both implemented following \cite{pan2021variational}.

\subsection{Per-category comparison}
In Table \ref{tab:supp_mvp} and Table \ref{tab:supp_crn}, we show the per-category evaluation results on MVP dataset and CRN dataset and compare our method with existing methods. It can be found that our method win on most categories and we achieve the best performance on the mean CD and mean F1-score metrics. This again  demonstrates that our method is superior to all baselines, benefited from the semantic prior distillation and self-supervised completion.

\subsection{More ablation study}
\noindent \textbf{Impact of Loss Function:} In our loss function, two kinds of losses are very important to the  training process: the point cloud similarity loss ($L_{cd,paired}$ and $L_{cd,unpaired}$ in our implementation) and latent code distillation loss ($L_{z,paired}$ and $L_{z,unpaired}$ in our implementation). In Table \ref{tab:impact_loss}, we show the performance of using some other alternatives (i.e., Earth Mover's Distance, Cosine Distance, Jensen–Shannon divergence, L1 distance) and different combinations.  Among all alternatives and  combinations, we find that using chamfer distance as the point cloud similarity loss and using Kullback-Leibler divergence as latent code distillation loss exhibits the best performance. 

\noindent \textbf{Competitors' Performance of utilizing Unpaired Incomplete Point Clouds:} In our work, we propose a self-supervised  completion module to better make use of the unpaired incomplete point clouds to train RaPD. While in the previous comparison, all competitors are trained without utilizing these data. To verify if these methods can also benefit from these data without adopting our prior distillation stage, we conduct experiments in Table \ref{tab:competitor_self}. We train PCN and TopNet by adding these unpaired incomplete training data. The same degradation and loss are used for them except for the prior distillation. It can be found in Table \ref{tab:competitor_self} that both competitors fail to take advantage of these data. The reason may be that without distilling the semantic prior, the network would  be severely suffered from over-fitting, hence the network would prefer to predict an incomplete point cloud rather than a complete one.

\subsection{More qualitative comparison} 
\begin{figure*}[h]
  \centering
  \includegraphics[width=\linewidth]{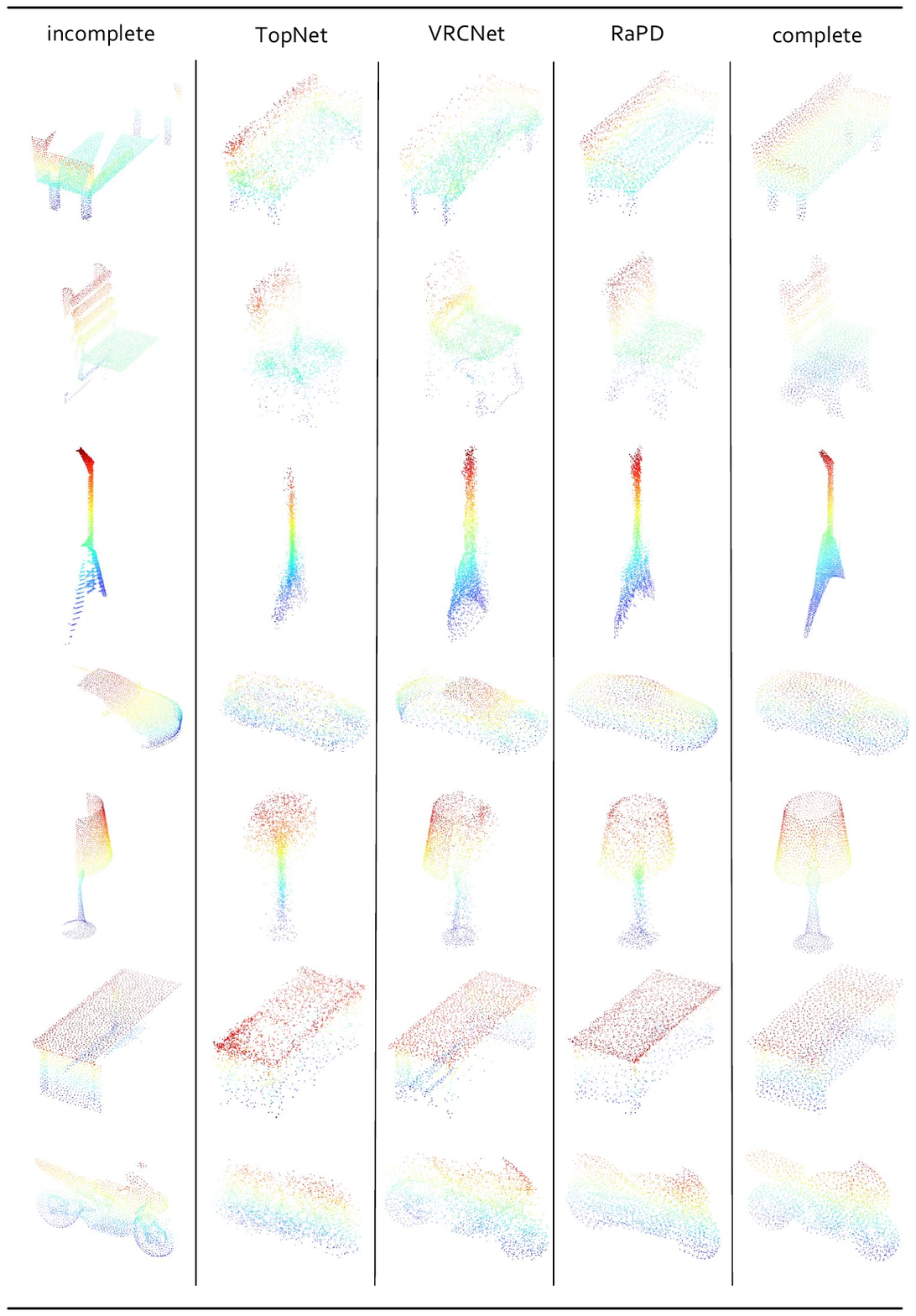}
  \caption{More qualitative comparison on MVP dataset. }
  \label{fig:mvp_supp}
\end{figure*}

\begin{figure*}[h]
  \centering
  \includegraphics[width=\linewidth]{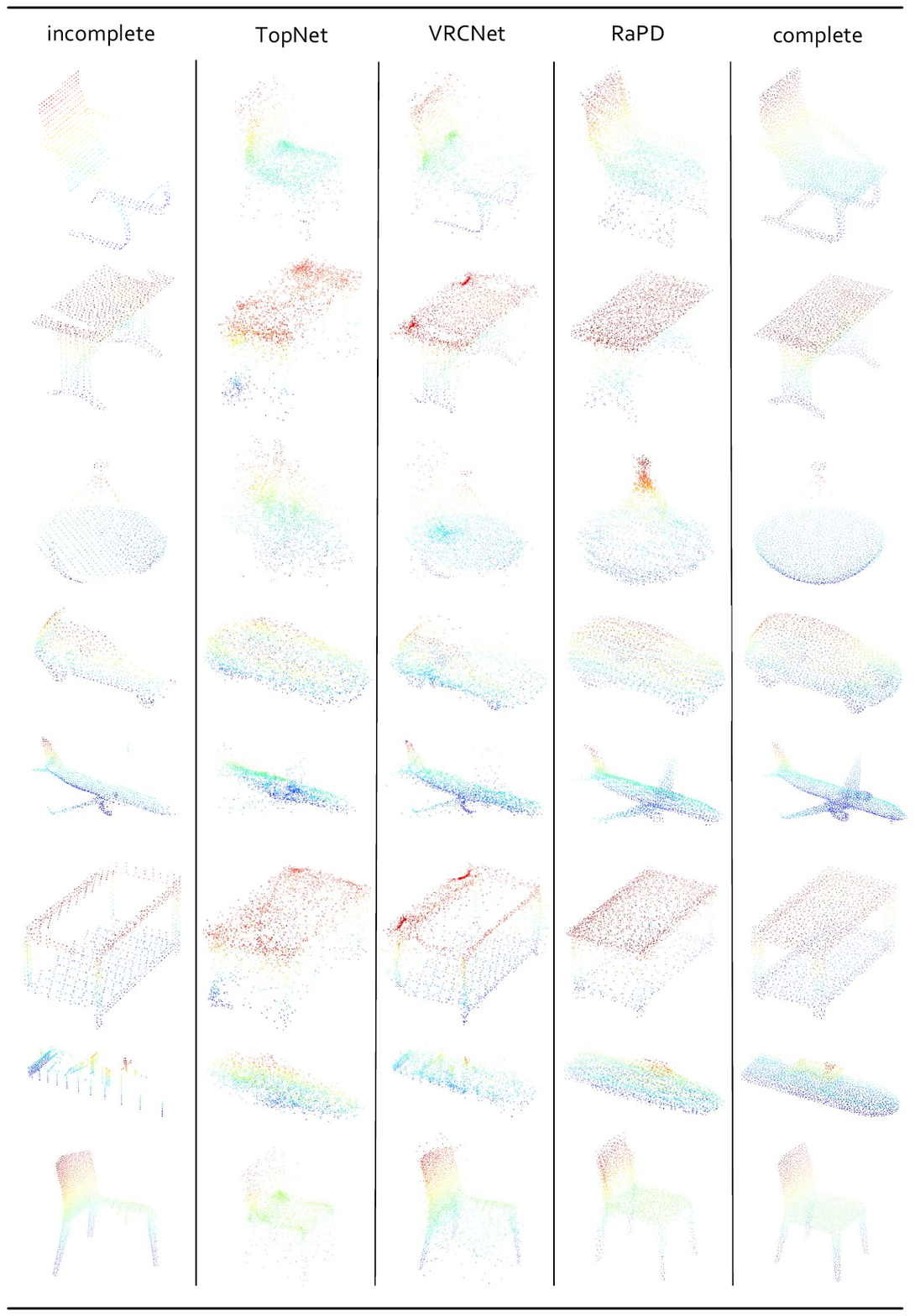}
  \caption{More qualitative comparison on CRN dataset. }
  \label{fig:crn_supp}
\end{figure*}

To provide the readers a wider view to better understand our method. We show more visualization results in Fig. \ref{fig:mvp_supp} and Fig. \ref{fig:crn_supp} for the MVP dataset and CRN dataset respectively. These results can demonstrate the good performance of our method on both homologous and heterologous experimental settings. We hope these visualizations can provide cues to benefit our readers' feature research to further make contribution to the field of semi-supervised point cloud completion.

\begin{table}
\begin{tabular}{cc|cc|cc}
\hline
     \multirow{2}*{PCS}      &    \multirow{2}*{LCD}        & \multicolumn{ 2}{|c|}{RaPD*} & \multicolumn{ 2}{c}{RaPD} \\
\cline{3-6}
        &         &         CD &   F1-score &         CD &   F1-score \\
\hline
        CD &        N/A &      11.61 &      0.299 &      43.86 &      0.109 \\

        CD & cosine distance &      11.51 &      0.301 &    13.12 &      0.285 \\

        CD & L1 distance &      11.45 &      0.302 &      11.73 &      0.298 \\

        CD & JS  distance &      11.43 &      0.301 &      10.12 &      0.326 \\

       EMD & KL distance &      11.37 &      0.301 &      11.16 &      0.307 \\

        CD & KL distance & {\bf 10.90} & {\bf 0.315} & {\bf 10.07} & {\bf 0.328} \\
\hline
\end{tabular}  

\caption{Ablation study about the impact of the loss function. PCS refers to point cloud similarity loss, and LCD refers to latent code distillation loss.}
\label{tab:impact_loss}

\end{table}

\begin{table}
\centering
\begin{tabular}{c|cc}
\hline
           &         CD &   F1-score \\
\hline
    TopNet \cite{tchapmi2019topnet} &     658.14 &      0.001 \\

       PCN \cite{yuan2018pcn} &     673.35 &      0.028 \\

      RaPD & {\bf 10.07} & {\bf 0.328} \\
\hline
\end{tabular}  

\caption{Ablation study about the performance of competitor utilizing unpaired incomplete point clouds.}
\label{tab:competitor_self}

\end{table}

\begin{table*}
\begin{tabular}{c|cc|cc|cc|cc|cc}
\hline
\multicolumn{ 1}{c|}{} & \multicolumn{ 2}{|c}{TopNet} & \multicolumn{ 2}{|c}{PCN} & \multicolumn{ 2}{|c}{VRCNet} & \multicolumn{ 2}{|c}{RaPD*} & \multicolumn{ 2}{|c}{RaPD} \\
\cline{2-11}
\multicolumn{ 1}{c|}{} &         CD &   F1-score &         CD &   F1-score &         CD &   F1-score &         CD &   F1-score &         CD &   F1-score \\
\hline
  airplane &       6.41 &      0.508 &       4.97 &      0.614 &       4.63 &      0.649 &       4.33 &      0.675 &  {\bf 3.70} & {\bf 0.687} \\

   cabinet &      14.12 &      0.126 &      12.45 &      0.418 &      15.63 &      0.155 & {\bf 10.44} &      0.187 &      10.91 & {\bf 0.199} \\

       car &       9.51 &      0.164 &       8.53 &      0.182 &       9.16 &      0.216 & {\bf 7.07} &      0.233 &       7.44 & {\bf 0.237} \\

     chair &      17.96 &      0.145 &      15.99 &      0.157 &      18.82 &      0.204 &      14.28 &      0.184 & {\bf 13.25} & {\bf 0.208} \\

      lamp &      27.01 &       0.180 &      22.89 &      0.217 &      20.55 & {\bf 0.345} &      19.94 &      0.258 & {\bf 15.72} &      0.285 \\

      sofa &      16.16 &      0.119 &      13.14 &      0.134 &      15.54 &      0.171 &      11.38 &       0.170 & {\bf 11.12} & {\bf 0.199} \\

     table &      16.63 &      0.221 &      15.59 &      0.228 &      16.94 & {\bf 0.327} &      14.16 &       0.260 & {\bf 13.80} &      0.279 \\

watercraft &      11.24 &       0.270 &      11.02 &       0.230 &       10.60 & {\bf 0.37} &        9.70 &      0.328 & {\bf 8.65} &      0.349 \\

       bed &      27.37 &      0.099 &       24.30 &      0.109 &      24.72 & {\bf 0.186} &      22.02 &      0.138 & {\bf 19.53} &      0.168 \\

     bench &      13.28 &      0.295 &      11.67 &      0.319 &       12.80 &      0.382 &       10.10 &      0.385 &  {\bf 9.20} & {\bf 0.392} \\

 bookshelf &      17.59 &      0.132 &      15.61 &      0.149 &      18.83 & {\bf 0.197} &      15.74 &      0.173 & {\bf 14.5} &       0.180 \\

       bus &      11.28 &      0.227 &       9.31 &       0.280 &       8.93 &      0.307 &       6.72 & {\bf 0.394} & {\bf 6.45} &      0.387 \\

    guitar &       4.38 &      0.519 &       2.68 &       0.650 &       4.69 &      0.532 & {\bf 2.08} & {\bf 0.754} &       2.27 &      0.719 \\

 motorbike &       8.94 &      0.248 &       8.12 &      0.255 &       8.13 & {\bf 0.313} &       6.87 &      0.306 & {\bf 6.75} &      0.291 \\

    pistol &      13.34 &      0.345 &       7.81 &      0.424 &       8.06 &      0.431 &       6.69 &       0.450 & {\bf 6.45} & {\bf 0.456} \\

skateboard &       6.52 &      0.472 &       8.05 &      0.543 &       5.74 &       0.560 &        4.80 & {\bf 0.606} & {\bf 3.78} &       0.590 \\

      mean &      14.37 &      0.236 &      12.54 &      0.271 &      13.36 &      0.319 &       10.90 &      0.315 & {\bf 10.07} & {\bf 0.328} \\
\hline
\end{tabular}  
\caption{Per-category results on MVP dataset. The model is trained using  5\%  paired point clouds of the MVP training set. CD loss is multiplied by $10^4$.}
\label{tab:supp_mvp}

\end{table*}

\begin{table*}
\begin{tabular}{c|cc|cc|cc|cc|cc}
\hline
\multicolumn{ 1}{c|}{} & \multicolumn{ 2}{|c}{TopNet} & \multicolumn{ 2}{|c}{PCN} & \multicolumn{ 2}{|c}{VRCNet} & \multicolumn{ 2}{|c}{RaPD*} & \multicolumn{ 2}{|c}{RaPD} \\
\cline{2-11}
\multicolumn{ 1}{c|}{} &         CD &   F1-score &         CD &   F1-score &         CD &   F1-score &         CD &   F1-score &         CD &   F1-score \\
\hline
  airplane &      10.97 &      0.421 &      10.41 &      0.469 &       9.42 &       0.470 &       5.97 &      0.611 & {\bf 4.89} & {\bf 0.605} \\

   cabinet &      26.04 &      0.073 &      23.28 &      0.089 &      27.24 &      0.098 &      19.45 &        0.100 & {\bf 18.86} & {\bf 0.119} \\

       car &      11.02 &      0.151 &      10.47 &      0.164 &      13.17 &      0.136 & {\bf 8.98} & {\bf 0.193} &      10.19 &      0.173 \\

     chair &      21.32 &      0.127 &      18.92 &      0.143 &      20.11 &      0.147 &      17.03 &      0.158 & {\bf 15.07} & {\bf 0.182} \\

      lamp &      37.69 &      0.131 &      28.86 &      0.145 &      25.64 & {\bf 0.275} &      24.02 &      0.197 & {\bf 19.10} &      0.236 \\

      sofa &      25.94 &      0.096 &      20.64 &      0.103 &       22.40 &      0.092 &      21.82 &      0.111 & {\bf 17.63} & {\bf 0.127} \\

     table &      25.49 &      0.148 &      19.58 &      0.207 &       18.60 &      0.172 &       17.90 &      0.202 &   {\bf 16.00} & {\bf 0.224} \\

watercraft &      16.94 &       0.210 &      13.72 &       0.230 &      11.65 & {\bf 0.307} &      12.77 &      0.262 & {\bf 10.28} &      0.285 \\

   mean &      22.05 &      0.171 &      18.34 &      0.195 &      18.56 &      0.215 &      15.98 &      0.229 & {\bf 13.96} & {\bf 0.244} \\
\hline
\end{tabular}  
\caption{Per-category results on CRN dataset. The training stage 1  is trained using  5\%  paired point clouds of the MVP training set. The training stage 2 is trained  using  5\%  paired point clouds of the CRN training set. CD loss is multiplied by $10^4$. }
\label{tab:supp_crn}

\end{table*}

\bibliographystyle{named}
\bibliography{ijcai23}

\end{document}